# Mix-nets: Factored Mixtures of Gaussians in Bayesian Networks with Mixed Continuous And Discrete Variables

Scott Davies and Andrew Moore
School of Computer Science
Carnegie Mellon University
Pittsburgh, PA 15213
[scottd, awm]@cs.cmu.edu

## Abstract

Recently developed techniques have made it possible to quickly learn accurate probability density functions from data in low-dimensional continuous spaces. In particular, mixtures of Gaussians can be fitted to data very quickly using an accelerated EM algorithm that employs multiresolution $k$d-trees (Moore, 1999). In this paper, we propose a kind of Bayesian network in which low-dimensional mixtures of Gaussians over different subsets of the domain's variables are combined into a coherent joint probability model over the entire domain. The network is also capable of modeling complex dependencies between discrete variables and continuous variables without requiring discretization of the continuous variables. We present efficient heuristic algorithms for automatically learning these networks from data, and perform comparative experiments illustrating how well these networks model real scientific data and synthetic data. We also briefly discuss some possible improvements to the networks, as well as possible applications.

## 1 INTRODUCTION

Bayesian networks are a popular method for representing joint probability distributions over many variables. A Bayesian network contains a directed acyclic graph $G$ with one vertex $V_i$ in the graph for each variable $X_i$ in the domain. The directed edges in the graph specify a set of independence relationships between the variables. Define $\vec{\Pi}_i$ to be the set of variables whose nodes in the graph are "parents" of $V_i$. The set of independence relationships specified by $G$ is then as follows: given the values of $\vec{\Pi}_i$ but no other information, $X_i$ is conditionally independent of all variables corresponding to nodes that are not $V_i$'s descendants in the graph. These independence relationships allows us to decompose the joint probability distribution $P(\vec{X})$ as $P(\vec{X}) = \prod_{i=1}^{N} P(X_i|\vec{\Pi}_i)$, where $N$ is the number of variables in the domain. Thus, if in addition to $G$ we also specify $P(X_i|\vec{\Pi}_i)$ for every variable $X_i$, then we have specified a valid probability distribution $P(\vec{X})$ over the entire domain.

Bayesian networks are most commonly used in situations where all the variables are discrete; if continuous variables are modeled at all, they are typically assumed to follow simple parametric distributions such as Gaussians. Some researchers have recently investigated the use of complex continuous distributions within Bayesian networks; for example, weighted sums of Gaussians have been used to approximate conditional probability density functions (Driver and Morrell, 1995). Such complex distributions over continuous variables are usually quite computationally expensive to learn. This expense may not be too problematic if an appropriate Bayesian network structure is known beforehand. On the other hand, if the dependencies between variables are not known *a priori* and the structure must be learned from data, then the number of conditional distributions that must be learned and tested while a structure-learning algorithm searches for a good network can become unmanageably large.

However, very fast algorithms for learning complex joint probability densities over small sets of continuous variables have recently been developed (Moore, 1999). This paper investigates how these algorithms can be used to learn Bayesian networks over many variables, each of which can be either continuous or discrete.

## 2 MIX-NETS

Suppose we have a very fast, black-box algorithm $A$ geared not towards finding accurate conditional models of the form $P_i(X_i|\vec{\Pi}_i)$, but rather towards finding accurate *joint* probability models $P_i(\vec{S}_i)$ over subsets of variables $\vec{S}_i$, such as $P_i(X_i, \vec{\Pi}_i)$. Furthermore, suppose it is only capable of generating joint models for relatively small subsets of the variables, and that the models returned for different sub-sets of variables are not necessarily consistent. Can we still combine many different models generated by $A$ into a valid probability distribution over the entire space?

Fortunately, the answer is yes, as long as the models returned by $A$ can be marginalized exactly. If for any given $P_i(X_i, \vec{\Pi}_i)$ we can compute a marginal distribution $P_i(\vec{\Pi}_i)$ that is consistent with it, then we can employ $P_i$ as a conditional distribution $P_i(X_i|\vec{\Pi}_i) = P_i(X_i, \vec{\Pi}_i)/P_i(\vec{\Pi}_i)$. In this case, given a directed acyclic graph $G$ specifying a Bayesian network structure over $N$ variables, we can simply use $A$ to acquire $N$ models $P_i(X_i, \vec{\Pi}_i)$, marginalize these models, and string them together to form a distribution over the entire space:

$$P(\vec{X}) = \prod_{i=1}^{N} (P_i(X_i, \vec{\Pi}_i)/P_i(\vec{\Pi}_i))$$

Even though the marginals of different $P_i$'s may be inconsistent with each other, the $P_i$'s are only *used* conditionally, and in a manner that prevents these inconsistencies from actually causing the overall model to become inconsistent. Of course, the fact that there are inconsistencies at all — suppressed or not — means that there is a certain amount of redundancy in the overall model. However, if allowing such redundancy lets us employ a particularly fast and effective model-learning algorithm $A$, it may be worth it.

Joint models over subsets of variables have been similarly conditionalized in previous research in order to use them within Bayesian networks. For example, the conditional distribution of each variable in the network given its parents can be modeled by conditionalizing another "embedded" Bayesian network that specifies the joint between the variable and its parents (Heckerman and Meek, 1997a). (Some theoretical issues concerning the interdependence of parameters in such models appear in (Heckerman and Meek, 1997a) and (Heckerman and Meek, 1997b).) Joint distributions formed by convolving a Gaussian kernel function with each of the datapoints have also been conditionalized for use in Bayesian networks over continuous variables (Hofmann and Tresp, 1995).

## 2.1 HANDLING CONTINUOUS VARIABLES

Suppose for the moment that $\vec{X}$ contains only continuous values. What sorts of models might we want $A$ to return? One powerful type of model for representing probability density functions over small sets of variables is a *Gaussian mixture model* (see e.g. (Duda and Hart, 1973)). Let $\vec{s}_j$ represent the values that the $j^{th}$ datapoint in the dataset $D$ assigns to a variable set of interest $\vec{S}$. In a Gaussian mixture model over $\vec{S}$, we assume that the data are generated independently through the following process: for each $\vec{s}_j$ in turn, nature begins by randomly picking a class, $c_k$, from a discrete set of classes $c_1, \ldots, c_M$. Then nature draws $\vec{s}_j$ from a multidimensional Gaussian whose mean vector $\vec{\mu}_k$ and covariance matrix $\Sigma_k$ depend on the class. This produces a distribution of the following mathematical form:

$$P(\vec{S}|\vec{\theta}) = \sum_{k=1}^{M} \alpha_k (2\pi||\Sigma_k||)^{-\frac{1}{2}} exp(-\frac{1}{2}(\vec{S}-\vec{\mu}_k)^T \Sigma_k^{-1} (\vec{S}-\vec{\mu}_k))$$

where $\alpha_k$ represents the probability of a point coming from the $k^{th}$ class, and $\vec{\theta}$ denotes the entire set of the mixture's parameters (the $\alpha$'s, $\mu$'s and $\Sigma$'s). It is possible to model any continuous probability distribution with arbitrary accuracy by using a mixture with a sufficiently large $M$.

Given a Gaussian mixture model $P_i(X_i, \vec{\Pi}_i)$, it is easy to compute the marginalization $P_i(\vec{\Pi}_i)$: the marginal mixture has the same number of Gaussians as the original mixture, with the same $\alpha$'s. The means and covariances of the marginal mixture are simply the means and covariances of the original mixture with all elements corresponding to the variable $X_i$ removed. Thus, Gaussian mixture models are suitable for combining into global joint probability density functions using the methodology described in section 2, assuming all variables in the domain are continuous. This is the class of models we employ for continuous variables in this paper, although many other classes may be used in an analogous fashion.

The functional form of the conditional distribution we use is similar to that employed in previous research by conditionalizing a joint distribution formed by convolving a Gaussian kernel function with all the datapoints (Hofmann and Tresp, 1995). The differences are that our distributions use fewer Gaussians, but these Gaussians have varying weights and varying non-diagonal covariance matrices. The use of fewer Gaussians makes our method more suitable for some applications such as compression, and may speed up inference. Our method may also yield more accurate models in many situations, but we have yet to verify this experimentally.

### 2.1.1 Learning Gaussian Mixtures From Data

The EM algorithm is a popular method for learning mixture models from data (see, e.g., (Dempster et al., 1977)). The algorithm is an iterative algorithm with two steps per iteration. The Expectation or "E" step calculates the distribution over the unobserved mixture component variables, using the current estimates for the model's parameters. The Maximization or "M" step then re-estimates the model's parameters to maximize the likelihood of both the observed data and the unobserved mixture component variables, assuming the distribution over mixture components calculated in the previous "E" step is correct. Each iteration of the EM algorithm increases the likelihood of the observed data or leaves it unchanged; if it leaves it unchanged, this usually indicates that the likelihood is at a local maximum. Unfortunately, each iteration of the basic algorithm described above is slow, since it requires a entire pass through the dataset. Instead, we use an accelerated EM algorithm in which multiresolution $k$d-trees (Moore et al., 1997) are used to dramatically reduce the computational cost of each iteration (Moore, 1999). We refer the interested reader to this previous paper (Moore, 1999) for details.

An important remaining issue is how to choose the appropriate number of Gaussians, $M$, for the mixture. If we restrict ourselves to too few Gaussians, we will fail to model the data accurately; on the other hand, if we allow ourselves too many, then we may "overfit" the data and our model may generalize poorly. A popular way of dealing with this tradeoff is to choose the model maximizing a scoring function that includes penalty terms related to the number of parameters in the model. We employ the Bayesian Information Criterion (Schwarz, 1978), or BIC, to choose between mixtures with different numbers of Gaussians. The BIC score for a probability model $P'(\vec{S})$ is as follows:

$$BIC(P') = \log P'(D_S) - \frac{\log R}{2}|P'|$$

where $D_S$ is the dataset $D$ restricted to the variables of interest $\vec{S}$, $R$ is the number of datapoints in the dataset, and $|P'|$ is the number of parameters in $P'$.

Rather than re-running the EM algorithm to convergence for many different choices of $M$ and choosing the resulting mixture that maximizes the BIC score, we use a heuristic algorithm that starts with a small number of Gaussians and stochastically tries adding or deleting Gaussians as the EM algorithm progresses. The details of this algorithm are described in a separate paper (Sand and Moore, 2000).

## 2.2 HANDLING DISCRETE VARIABLES

Suppose now that a set of variables $\vec{S}_i$ we wish to model includes discrete variables as well as continuous variables. Let $\vec{Q}_i$ be the discrete variables in $\vec{S}_i$, and $\vec{C}_i$ the continuous variables in $\vec{S}_i$. One simple model for $P_i(\vec{Q}_i, \vec{C}_i)$ is a lookup table with an entry for each possible set $\vec{q}_i$ of assignments to $\vec{Q}_i$. The entry in the table corresponding to a particular $\vec{q}_i$ contains two things: the marginal probability $P_i(\vec{q}_i)$, and a Gaussian mixture modeling the conditional distribution $P_i(\vec{C}_i|\vec{q}_i)$. Let us refer to tables of this form as *mixture tables*. We obtain the mixture table's estimate for each $P_i(\vec{q}_i)$ directly from the data: it is simply the fraction of the records in the dataset that assigns the values $\vec{q}_i$ to $\vec{Q}_i$. Given an algorithm $A$ for learning Gaussian mixtures from continuous data, we use it to generate each conditional distribution $P_i(\vec{C}_i|\vec{q}_i)$ in the mixture table by calling it with the subset of the dataset $D$ that is consistent with the values specified by $\vec{q}_i$.

Suppose now that we are given a network structure over the entire set of variables, and for each variable $X_i$ we are given a mixture table for $P_i(\vec{S}_i) = P_i(X_i, \vec{\Pi}_i)$. We must now calculate new mixture tables for the marginal distributions $P_i(\vec{\Pi}_i)$ so that we can use them for the conditional distributions $P_i(X_i|\vec{\Pi}_i) = P_i(X_i, \vec{\Pi}_i)/P_i(\vec{\Pi}_i)$. Let $\vec{C}_i$ represent the continuous variables in $\{X_i\} \cup \vec{\Pi}_i$; $\vec{Q}_i$ represent the discrete variables in $\{X_i\} \cup \vec{\Pi}_i$; $\vec{C}_{\Pi_i}$ represent the continuous variables in $\vec{\Pi}_i$; and $\vec{Q}_{\Pi_i}$ represent the discrete variables in $\vec{\Pi}_i$. (Either $\vec{Q}_{\Pi_i} = \vec{Q}_i$ or $\vec{C}_{\Pi_i} = \vec{C}_i$, depending on whether $X_i$ is continuous or discrete.)

If $X_i$ is continuous, then the marginalized mixture table for $P_i(\vec{\Pi}_i)$ has the same number of entries as the original table for $P_i(X_i, \vec{\Pi}_i)$, and the estimates for $P(\vec{Q}_i)$ in the marginalized table are the same as in the original table. For each combination of assignments to $\vec{Q}_i$, we marginalize the appropriate Gaussian mixture $P_i(\vec{C}_i|\vec{Q}_i) = P_i(\vec{C}_i|\vec{Q}_{\Pi_i})$ in the original table to a new mixture $P_i(\vec{C}_{\Pi_i}|\vec{Q}_{\Pi_i})$, and use this new mixture in the corresponding spot in the marginalized table.

If $X_i$ is discrete, then for each combination of values for $\vec{Q}_{\Pi_i}$ we combine several different Gaussian mixtures for various $P_i(\vec{C}_{\Pi_i}|\vec{Q}_i)$'s into a new Gaussian mixture for $P_i(\vec{C}_{\Pi_i}|\vec{Q}_{\Pi_i})$. First, the values of $P_i(\vec{Q}_{\Pi_i})$ in the marginalized table are computed from the original table as $P_i(\vec{Q}_{\Pi_i}) = \sum_{X_i} P_i(X_i, \vec{Q}_{\Pi_i})$. $P_i(X_i|\vec{Q}_{\Pi_i})$ is then calculated as $P_i(X_i, \vec{Q}_{\Pi_i})/P_i(\vec{Q}_{\Pi_i})$. Finally, we combine the Gaussian mixtures corresponding to different values of $X_i$ according to the relationship

$$P_i(\vec{C}_{\Pi_i}|\vec{Q}_{\Pi_i}) = \sum_{X_i} P_i(X_i|\vec{Q}_{\Pi_i})P_i(\vec{C}_{\Pi_i}|\vec{Q}_i).$$

We have now described the steps necessary to use mixture tables in order to parameterize Bayesian networks over domains with both discrete and continuous variables. Note that mixture tables are not particularly well-suited for dealing with discrete variables that can take on many possible values, or for scenarios involving many dependent discrete variables — in such situations, the continuous data will be shattered into many separate Gaussian mixtures, each of which will have little support. Better ways of dealing with discrete variables are undoubtedly possible, but we leave them for future research (see section 6.3).

## 3 LEARNING MIX-NET STRUCTURES

Given a Bayesian network structure over a domain with both discrete and continuous variables, we now know how to learn mixture tables describing the joint probability of each variable and its parent variables, and how to marginalize these mixture tables to obtain the conditional distributions needed to compute a coherent probability function over the entire domain. But what if we don't know *a priori* what dependencies exist between the variables in the domain — can we learn these dependencies automatically and find the best Bayesian network structure on our own, or at least find a "good" network structure?

In general, finding the optimal Bayesian network structure with which to model a given dataset is NP-complete (Chickering, 1996), even when all the data is discrete and there are no missing values or hidden variables. A popular heuristic approach to finding networks that model discrete data well is to hillclimb over network structures, using a scoring function such as the BIC as the criterion to maximize. (See, e.g., (Heckerman et al., 1995).) Unfortunately, hillclimbing usually requires scoring a very large number of networks. While our algorithm for learning Gaussian mixtures from data is comparatively fast for the complex task it performs, it is still too expensive to re-run on the hundreds of thousands of different variable subsets that would be necessary to parameterize all the networks tested over an extensive hillclimbing run. (Such a hillclimbing algorithm has previously been used to find Bayesian networks suitable for modeling continuous data with complex distributions (Hofmann and Tresp, 1995), but in practice this method is restricted to datasets with relatively small numbers of variables and datapoints.)

However, there are other heuristic algorithms that often find networks very close in quality to those found by hillclimbing but with much less computation. A frequently used class of algorithms involves measuring all pairwise interactions between the variables, and then constructing a network that models the strongest of these pairwise interactions (e.g. (Chow and Liu, 1968; Sahami, 1996; Friedman et al., 1999)). We use such an algorithm in this paper to automatically learn the structures of our Bayesian networks.

In order to measure the pairwise interactions between the variables, we start with an empty Bayesian network $B_\epsilon$ in which there no arcs — i.e., in which all variables are assumed to be independent. We use our mixture-learning algorithm to calculate $B_\epsilon$'s parameters, and then calculate $B_\epsilon$'s BIC score. (The number of parameters in the network is computed as the sum of the parameters in each network node, where the parameters of a node for variable $X_i$ are the parameters of $P_i(X_i, \Pi)$.) We then calculate the BIC score of every possible Bayesian network containing exactly one arc. With $N$ variables, there are $O(N^2)$ such networks. Let $B_{ij}$ denote the network with a single arc from $X_i$ to $X_j$. Note that to compute the BIC score of $B_{ij}$, we need not recompute the mixture tables for any variable other than $X_j$, since the others can simply be copied from $B_\epsilon$. Now, define $I(X_i, X_j)$, the "importance" of the dependency between variable $X_i$ and $X_j$, as follows:

$$I(X_i, X_j) = BIC(B_{ij}) - BIC(B_\epsilon).$$

After computing all the $I(X_i, X_j)$'s, we initialize our current working network $B$ to the empty network $B_\epsilon$, and then greedily add arcs to $B$ using the $I(X_i, X_j)$'s as hints for what arcs to try adding next. At any given point in the algorithm, the set of variables is split into two mutually exclusive subsets, DONE and PENDING. All variables begin

- $B := B_\epsilon$, PENDING := the set of all variables, DONE := {}
- While there are still variables in PENDING:
  - Consider all pairs of variables $X_d$ and $X_p$ such that $X_d$ is in DONE and $X_p$ is in PENDING. Of these, let $X_d^{max}$ and $X_p^{max}$ be the pair of variables that maximizes $I(X_d, X_p)$. Our algorithm selects $X_p^{max}$ as the next variable to consider adding arcs to. (Ties are handled arbitrarily, as is the case where DONE is currently empty.)
  - Let $K' = \min(K, |DONE|)$, where $K$ is a user-defined parameter. Let $X_d^1, X_d^2, \ldots X_d^{K'}$ denote the $K'$ variables in DONE with the highest values of $I(X_d^i, X_p^{max})$, in descending order of $I(X_d^i, X_p^{max})$.
  - For $i = 1$ to $K'$:
    * If $X_p^{max}$ now has MAXPARS parents in $B$, or if $I(X_d^i, X_p^{max})$ is less than zero, break out of the for loop over $i$ and do not consider adding any more parents to $X_p^{max}$.
    * Let $B'$ be a network identical to $B$ except with an additional arc from $X_d^i$ to $X_p^{max}$. Call our mixture-learning algorithm to update the parameters for $X_p^{max}$'s node in $B'$, and compute $BIC(B')$.
    * If $BIC(B') > BIC(B), B := B'$.
  - Move $X_p^{max}$ from PENDING to DONE.

Figure 1: Our network structure learning algorithm.

in the PENDING set. Our algorithm proceeds by selecting a variable in the PENDING set, adding arcs to that variable from other variables in the DONE set, moving the variable to the DONE set, and repeating until all variables are in DONE. High-level pseudo-code for the algorithm appears in Figure 1.

The algorithm generates $O(N^2)$ mixture tables containing two variables each in order to calculate all the pairwise dependency strengths $I(X_i, X_j)$, and then $O(N*K)$ more tables containing $MAXPARS + 1$ or fewer variables each during the greedy network construction. $K$ is a user-defined parameter that determines the maximum number of potential parents evaluated for each variable during the greedy network construction.

If MAXPARS is set to 1 and $I(X_i, X_j)$ is symmetric, then our heuristic algorithm reduces to a maximum spanning tree algorithm (or to a maximum-weight forest algorithm if some of the $I$'s are negative). Out of all possible Bayesian networks in which each variable has at most one parent, this maximum spanning tree is the Bayesian network $B^1_{opt}$ that maximizes the scoring function. (This is a trivial generalization of the well-known algorithm (Chow and Liu, 1968) for the case where the unpenalized log-likelihood is the objective criteria being maximized.) If MAXPARS is set above 1, our heuristic algorithm will always model a superset of the dependencies in $B^1_{opt}$, and will always find a network with at least as high a $BIC$ score as $B^1_{opt}$'s.

There are a few details that prevent our $I(X_i, X_j)$'s from being perfectly symmetric. (See (Davies and Moore, 2000).) However, $I$ is close enough to symmetric that it's often worth simply assuming that it is symmetric, since this cuts down the number of calls we need to make to our mixture-learning algorithm in order to calculate the $I(X_i, X_j)$'s by roughly a factor of 2.

Since learning joint distributions involving real variables is expensive, calling our mixture table generator even just $O(N^2)$ times to measure all of the $I(X_i, X_j)$'s can take a prohibitive amount of time. We note that the $I(X_i, X_j)$'s are only used to choose the order in which the algorithm selects variables to move from PENDING to DONE, and to select which arcs to try adding to the graph. The actual values of $I(X_i, X_j)$ are irrelevant — the only things that matter are their ranks and whether they are greater than zero. Thus, in order to reduce the expense of computing the $I(X_i, X_j)$'s, we can try computing them on a *discretized* version of the dataset rather than the original dataset that includes continuous values. The resulting ranks of $I(X_i, X_j)$ will not generally be the same as they would if they were computed from the original dataset, but we would expect them to be highly correlated in many practical circumstances.

Our structure-learning algorithm is similar to the "Limited Dependence Bayesian Classifiers" previously employed to learn networks for classification (Sahami, 1996), except that our networks have no special target variable, and we add the potential parents to a given node one at a time to ensure that each actually increases the network's score. Alternatively, our learning algorithm can be viewed as a restriction of the "Sparse Candidate" algorithm (Friedman et al., 1999) in which only one set of candidate parents is generated for each node, and in which the search over network structures restricted to those candidate parents is performed greedily. (We have also previously used a very similar algorithm for learning networks with which to compress discrete datasets (Davies and Moore, 1999).)

## 4 EXPERIMENTS

### 4.1 ALGORITHMS

We compare the performance of the network-learning algorithm described above to the performance of four other algorithms, each of which is designed to be similar to our network-learning algorithm except in one important respect. First we describe a few details about how our primary network-learning algorithm is used in our experiments, and then we describe the four alternative algorithms.

The **Mix-Net** algorithm is our primary network-learning algorithm. For our experiments on both datasets, we set MAXPARS to 3 and $K$ to 6. When generating any given Gaussian mixture, we give our accelerated EM algorithm thirty seconds to find the best mixture it can. In order to make the most of these thirty-second intervals, we also limit our overall training algorithm to using a sample of at most 10,000 datapoints from the training set. Rather than computing the $I(X_i, X_j)$'s with the original dataset, we compute them with a version of the dataset in which each continuous variable has been discretized to 16 different values. The boundaries of the 16 bins for each variable's discretization are chosen so that the numbers of datapoints in each bin are approximately equal.

Mixture tables containing many discrete variables (or a few discrete variables each of which can take on many values) can severely overfit data, since some combinations of the discrete variables may occur rarely in the data. For now, we attempt to address this problem by slightly smoothing the distributions represented in the tables. (See (Davies and Moore, 2000) for details.)

The **Independent Mixtures** algorithm is identical to our mix-net learning algorithm in almost all respects; the main difference is that here the MAXPARS parameter has been set to zero, thus forcing all variables to be modeled independently. We also give this algorithm more time to learn each individual Gaussian mixture, so that it is given a total amount of computational time at least as great as that used by our mix-net learning algorithm.

The **Trees** algorithm is identical to our mix-net algorithm in all respects except that the MAXPARS parameter has been set to one, and we give it more time to learn each individual Gaussian mixture (as we did for the Independent Mixtures algorithm).

The **Single-Gaussian** algorithm is identical to our primary network-learning algorithm except for the following differences. When learning a given Gaussian mixture $P_i(\vec{C_i}|\vec{Q_i})$, we use only a single multidimensional Gaussian. (Note, however, that some of the marginal distributions $P_i(\vec{C_{\Pi_i}}|\vec{Q_{\Pi_i}})$ may contain multiple Gaussians when the variable marginalized away is discrete.) Since single Gaussians are much easier to learn in high-dimensional spaces than mixtures are, we allow this single-Gaussian algorithm much more freedom in creating large mixtures. We set both MAXPARS and $K$ to the total number of variables in the domain minus one. We also allow the algorithm to use all datapoints in the training set rather than restrict it to a sample of 10,000. Finally, we use the original real-valued dataset rather than a discretized version of the dataset when computing the $I(X_i, X_j)$'s.

The **Pseudo-Discrete** algorithm is similar to our primary network-learning algorithm in that it uses the same sort of greedy algorithm to select which arcs to try adding to the network. However, the networks this algorithm produces do not employ Gaussian mixtures. Instead, the distributions it uses are closely related to the distributions that would be modeled by a Bayesian network for a completely discretized version of the dataset. For each continuous variable $X_i$ in the domain, we break $X_i$'s range into $F$ buckets. The boundaries of the buckets are chosen so that the number of datapoints lying within each bucket is approximately equal. The conditional distribution for $X_i$ is modeled with a table containing one entry for every combination of its parent variables, where each continuous parent variable's value is discretized according to the $F$ buckets we have selected for that parent variable. Each entry in the table contains a histogram for $X_i$ recording the conditional probability that $X_i$'s value lies within the boundaries of each of $X_i$'s $F$ buckets. We then translate the conditional probability associated with each bucket into a conditional probability density spread uniformly throughout the range of that bucket. (Discrete variables are handled in a similar manner, except the translation from conditional probabilities to conditional probability densities is not performed.)

When performing experiments with the Pseudo-Discrete network-learning algorithm, we re-run it for several different choices of $F$: 2, 4, 8, 16, 32, and 64. Of the resulting networks, we pick the one that maximizes the BIC. When the algorithm uses a particular value for $F$, the $I(X_i, X_j)'s$ are computed using a version of the dataset that has been discretized accordingly, and then arcs are added greedily as in our mix-net learning algorithm. The networks produced by this algorithm do not have redundant parameters as our mix-nets do, as each node contains only a model of its variable's conditional distribution given its parents rather than a joint distribution.

Note that there are better ways of discretizing real variables in Bayesian networks (e.g. (Kozlov and Koller, 1997; Monti and Cooper, 1998a)); the simple discretization algorithm discussed here and currently implemented for our experiments is certainly not state-of-the-art.

## 4.2 DATASETS AND RESULTS

We tested the previously described algorithms on two different datasets taken from real scientific experiments. The "Bio" dataset contains data from a high-throughput biological cell assay. There are 12,671 records and 31 variables. 26 of the variables are continuous; the other five are discrete. Each discrete variable can take on either two or three different possible values. The "Astro" dataset contains data taken from the Sloan Digital Sky Survey, an extensive astronomical survey currently in progress. This dataset contains 111,456 records and 68 variables. 65 of the variables are continuous; the other three are discrete, with arities ranging from three to 81. All continuous values are scaled to lie within $[0, 1]$ after a small amount of uniform noise is added to them to prevent potential problems caused by point distributions or deterministic relationships in the data.

For each dataset and each algorithm, we performed tenfold cross-validation, and recorded the log-likelihoods of the test sets given the resulting models. Figure 2 shows the mean log-likelihoods of the test sets according to models generated by our five network-learning algorithms, as well as the standard deviation of the means. (Note that the log-likelihoods are positive since most of the variables are continuous and bounded within $[0, 1]$, which implies that the models usually assign probability densities greater than one to regions of the space containing most of the datapoints. The probability distributions modeled by the networks are properly normalized, however.)

On the Bio dataset, our primary mix-net learner achieved significantly higher log-likelihood scores than the other four model learners. The fact that it significantly outperformed the independent mixture algorithm and the tree-learning algorithm indicates that it is effectively utilizing relationships between variables, and that it includes useful relationships more complex than mere pairwise dependencies. The fact that its networks outperformed the pseudo-discrete networks and the single-Gaussian networks indicates that the Gaussian mixture models used for the network nodes' parameterizations helped the networks achieve much better prediction than possible with simpler parameterizations. Our primary mix-net learning algorithm took about an hour and a half of CPU time on a 400 MHz Pentium II to generate its model for each of the ten cross-validation splits for this dataset.

The mix-net learner similarly outperformed the other algorithms on the Astro dataset. The algorithm took about three hours of CPU time to generate its model for each of the cross-validation splits for this dataset.

To verify that the mix-net's performance gain over the pseudo-discrete network's was due to its use of Gaussian mixtures rather than piecewise constant densities, we constructed a synthetic dataset from the Bio dataset. All real values in the original dataset were discretized in a manner identical to the manner in which the pseudo-discrete networks discretized them, with 16 buckets per variable. (Out of the many different numbers of buckets we tried with the pseudo-discrete networks, 16 was the number that worked best on the Bio dataset.) Each discretized value was then translated back into a real value by sampling it uniformly from the corresponding bucket's range. The resulting synthetic dataset is similar in many respects to the original dataset, but its probability densities are now composed of piecewise constant axis-aligned hyperboxes — precisely the kind of distributions that the pseudo-discrete networks model. The test-set performance of the pseudo-discrete networks on this synthetic dataset is the same as their test-set performance on the original dataset (59100+/-100). As one would expect, the test-set performance of mix-nets on this synthetic task is indeed worse (57600+/-200). However, it is not dramatically worse.

As a second synthetic task, we generated 12,671 samples from the network learned by the Independent Mixtures algorithm during one of it cross-validation runs on the Bio dataset. The test-set log-likelihood of the models learned by the Independent Mixtures algorithm on this dataset was 32580 +/- 60, while our primary mix-net learning algorithm scored a slightly worse 31960 +/- 80. However, the networks learned by the mix-net learning algorithm did not actually model any spurious dependencies between variables. The networks learned by the Independent Mixtures algorithm were better only because the Independent Mixtures algorithm was given more time to learn each of its Gaussian mixtures.

|  | Bio | Astro |
|---|---|---|
| Independent Mixtures | 33300 +/- 500 | 2746000 +/- 5000 |
| Single-Gaussian Mixtures | 65700 +/- 200 | 2436000 +/- 5000 |
| Pseudo-Discrete | 59100 +/- 100 | 3010000 +/- 1000 |
| Tree | 74600 +/- 300 | 3280000 +/- 8000 |
| Mix-Net | 80900 +/- 300 | 3329000 +/- 5000 |

Figure 2: Mean log-likelihoods (and the standard deviations of the means) of test sets in a 10-fold cross-validation.

## 5 POSSIBLE APPLICATIONS

Space restrictions prevent us from going into much detail on possible applications; for a more extended discussion, see the tech report version of this paper (Davies and Moore, 2000). The networks discussed here are trivially applicable to anomaly detection tasks in which all variables are observed. They may be well-suited to compressing datasets containing both discrete and real values (with the real values compressed lossily). It is also easy to modify our greedy network-learning algorithm to learn networks for classification tasks; the resulting networks would be similar in structure to those generated by previous algorithms (Friedman et al., 1997, 1998; Sahami, 1996), but with more flexible parameterizations.

While it is possible to perform exact inference in some kinds of networks modeling continuous values (e.g. (Driver and Morrell, 1995; Alag, 1996)), general-purpose exact inference in arbitrarily-structured mix-nets with continuous variables may not be possible. However, inference in these networks can be performed via stochastic sampling methods (such as importance sampling or Gibbs sampling). Other approximate inference methods such as variational inference or discretization-based inference are also worth investigating.

## 6 CONCLUSIONS / FUTURE RESEARCH

We have described a practical method for learning Bayesian networks capable of modeling complex interactions between many continuous and discrete variables, and have provided experimental results showing that the method is both feasible and effective on scientific data with dozens of variables. The networks learned by this algorithm and related algorithms show considerable potential for many important applications. However, there are many ways in which our method can be improved upon. We now briefly discuss a few of the more obvious possibilities for improvement.

### 6.1 VARIABLE GROUPING

The mixture tables in our network include a certain degree of redundancy, since the mixture table for each variable models the joint probability of that variable with its parents rather than just the conditional probability of that variable given its parents. One possible approach for reducing the amount of redundancy in the network is to allow each node to represent a *group* of variables, where each variable must be represented in exactly one group. See the technical report version of this paper (Davies and Moore, 2000) for a slightly more extensive discussion.

### 6.2 ALTERNATIVE STRUCTURE-LEARNERS

So far we have only developed and experimented with variations of one particular network structure-learning algorithm. There is a wide variety of structure-learning algorithms for discrete Bayesian networks (see, e.g., (Cooper and Herskovits, 1992; Lam and Bacchus, 1994; Heckerman et al., 1995; Friedman et al., 1999)), many of which could be employed when learning mix-nets. The quicker and dirtier of these algorithms might be applicable directly to learning mix-net structures. The more time-consuming algorithms such as hillclimbing can be used to learn Bayesian networks on discretized versions of the datasets; the resulting networks may then be used as hints for which sets of dependencies might be worth trying in a mix-net. Such approaches have previously been shown to work well on real datasets (Monti and Cooper, 1998b).

### 6.3 ALTERNATIVE PARAMETER-LEARNERS

While the accelerated EM algorithm we use to learn Gaussian mixtures is fast and accurate for low-dimensional mixtures, its effectiveness decreases dramatically as the number of variables in the mixture increases. This is the primary reason we have not yet attempted to learn mixture networks with more than four variables per mixture. Further research is currently being conducted on alternate data structures and algorithms which with to accelerate EM in the hopes that they will scale more gracefully to higher dimensions (Moore, 2000). In the meantime, it would be trivial to allow the use of some high-dimensional single-Gaussian mixtures within mix-nets as we do for the "competing" Single-Gaussian algorithm described in Section 4.1. Other methods for accelerating EM have also been developed in the past (e.g., (Neal and Hinton, 1998; Bradley et al., 1998)), some of which might be used in our Bayesian network-learning algorithm instead of or in addition to the accelerated EM algorithm employed in this paper.

Our current method of handling discrete variables does not deal well with discrete variables that can take on many possible values, or with combinations involving many discrete variables. Better methods of dealing with these situations are also grounds for further research. One approach would be to use mixture models in which the hidden class variable determining which Gaussian each datapoint's continuous values come from also determines distributions over the datapoint's discrete values, where each discrete value is assumed to be conditionally independent of the others given the class variable. Such an approach has been used previously in AutoClass (Cheeseman and Stutz, 1996). The accelerated EM algorithm exploited in this paper would have to be generalized to handle this class of models, however. Another approach would be to use decision trees over the discrete variables rather than full lookup tables, a technique previously explored for Bayesian networks over discrete domains (Friedman and Goldszmidt, 1996). In previous work (Koller et al., 1999) employing representations closely related to the those employed in this paper, a combination of these two approaches has been explored briefly in order to represent potentials in junction trees; further exploration along these lines may prove fruitful.

The Gaussian mixture learning algorithm we currently employ attempts to find a mixture maximizing the joint likelihood of all the variables in the mixture rather than a conditional likelihood. Since the mixtures are actually used to compute conditional probabilities, some of their representational power may be used inefficiently. The EM algorithm has recently been generalized to learn joint distributions specifically optimized for being used conditionally (Jebara and Pentland, 1999). If this modified EM algorithm can be accelerated in a manner similar to our current accelerated EM algorithm, it may result in significantly more accurate networks.

Finally, further comparisons with alternative methods for modeling distributions over continuous variables (such as the Gaussian kernel functions used in (Hofmann and Tresp, 1995)) are warranted.